\documentclass[10pt,twocolumn,letterpaper]{article}

\usepackage{wacv}
\usepackage{times}
\usepackage{epsfig}
\usepackage{graphicx}
\usepackage{amsmath}
\usepackage{amssymb}

\usepackage{soul}
\usepackage{verbatim}
\usepackage{subfigure}
\usepackage{calc}
\usepackage{amstext}
\usepackage{stackengine}
\newcommand\IncG[2][]{\addstackgap{%
\raisebox{-.5\height}{\includegraphics[#1]{#2}}}}
\usepackage{array}
\usepackage{amsmath,amssymb} 
\usepackage{xcolor}

\usepackage{helvet}  
\usepackage{courier}  
\usepackage{url}  

\wacvfinalcopy 

\def\wacvPaperID{****} 

\ifwacvfinal\fi
\begin{document}

\title{Revisiting Single Image Depth Estimation: \\
Toward Higher Resolution Maps
with Accurate Object Boundaries}

\author{ Junjie Hu$^{1,2}$ \hspace{1cm} Mete Ozay$^1$ \hspace{1cm} Yan Zhang$^{1,2}$ \hspace{1cm} Takayuki Okatani$^{1,2}$\\
$^1$ Graduate School of Information Sciences, Tohoku University, Japan \\
$^2$ Center for Advanced Intelligence project, RIKEN, Japan\\
{\tt\small \{junjie.hu, mozay, zhang, okatani\}@vision.is.tohoku.ac.jp}
}


\maketitle
\ifwacvfinal\thispagestyle{empty}\fi
\begin{abstract}
This paper considers the problem of single image depth estimation. 
The employment of convolutional neural networks (CNNs) has recently brought about significant advancements in the research of this problem. 
However, most existing methods suffer from loss of spatial resolution in the estimated depth maps; a typical symptom is distorted and blurry reconstruction of object boundaries.
In this paper, toward more accurate estimation with a focus on depth maps with higher spatial resolution, we propose two improvements to existing approaches.
One is about the strategy of fusing features extracted at different scales, for which we propose an improved network architecture consisting of four modules: an encoder, decoder, multi-scale feature fusion module, and refinement module.
The other is about loss functions for measuring inference errors used in training. We show that three loss terms, which measure errors in depth, gradients and surface normals, respectively, contribute to improvement of accuracy in an complementary fashion.
Experimental results show that these two improvements enable to attain higher accuracy than the current state-of-the-arts, which is given by finer resolution reconstruction, for example, with small objects and object boundaries. 
\end{abstract}

\section{Introduction}

\begin{figure}[t]
\centering
\subfigure[] {\includegraphics[height=28mm]{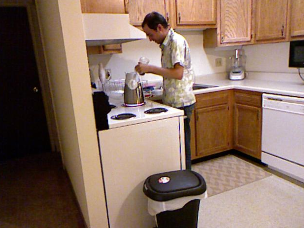}}
~\subfigure []{\includegraphics[height=28mm]{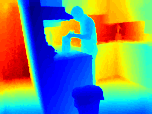}} \\
\subfigure []{\includegraphics[height=28mm]{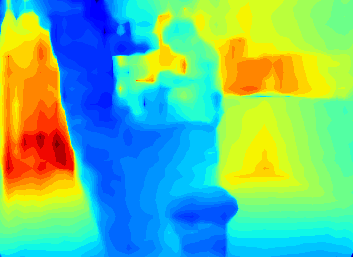}}
~\subfigure []{\includegraphics[height=28mm]{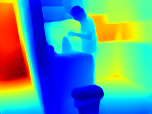}}
\caption{Example comparison of estimated depth maps; (a) RGB input, (b) ground truth depth, (c) the current state-of-the-art \cite{fu2018deep}, and (d) our method.}
\label{fig_one}
\end{figure}

The problem of estimating the depth map of a scene from its single image has attracted a lot of attention in the field of computer vision, since depth maps have a lot of applications, 
such as augmented reality \cite{lee2011depth}, human computer interaction \cite{Ren2011DepthCB}, human activity recognition \cite{Gupta2013HumanAR}, scene recognition \cite{zia2017rgb}, and segmentation \cite{Lu2018CurveStructureSF,qi2018geonet}. 
The recent employment of convolutional neural networks (CNNs) has accelerated the research of the problem \cite{Eigen2014depth,liu2015deep,Li2015DepthAS,cao2017estimating,chen2016single}. In early studies, depth maps can be estimated only with much lower resolution than input images, which is mostly attributable to a series of downsampling operations performed in CNNs. 
To recover the degraded spatial resolution, a number of studies have been conducted so far. First, a new up-sampling method called up-projection was proposed in \cite{laina2016deeper}. Integration of CRFs into CNNs for depth refinement, along with its end-to-end training method, was proposed in \cite{Xu2017MultiscaleCC}. Several works applied joint multi-task learning \cite{Eigen2015PredictingDS,Dharmasiri2017JointPO}, which provides a good amount of improvement of estimated depth maps.
Recently, dilated convolution was employed in \cite{fu2018deep}, updating the state-of-the-art accuracy. 

In this paper, we argue that  there is room for further improvements despite these previous efforts. 
In the depth maps estimated by the previous methods, including the one that has achieved the state-of-the-art accuracy, we can observe distortions in object shapes, missing small objects, and mosaic patterns, as shown in Fig.\ref{fig_one}. Depth maps with high spatial resolution, e.g., precise object boundaries, are especially important for some applications such as object recognition \cite{song2017depth} and depth-aware image re-rendering and editing \cite{li2017two}. 

The features learned in different layers of CNNs for depth estimation, which represent information at different scales, should contain different depth cues. For example, the lower layer features may have information on 
the details of object shapes, while the higher layer features represent global depth information without details of object shapes. 
Therefore, 
the former and latter should be used in a complementary fashion to estimate more accurate depth maps. 
Toward this end, we propose a feature fusion module that up-scales lower layer features to the same (high) level of resolution using skip connections, and then learns how to fuse the up-scaled different scale features with a convolutional layer. 
We also employ a decoder module to decode the high level feature extracted by the encoder.
Then, the outputs of the two modules are integrated and further refined to provide final estimation. 

\begin{figure*}[t]
\centering
\subfigure {\includegraphics[scale=0.35]{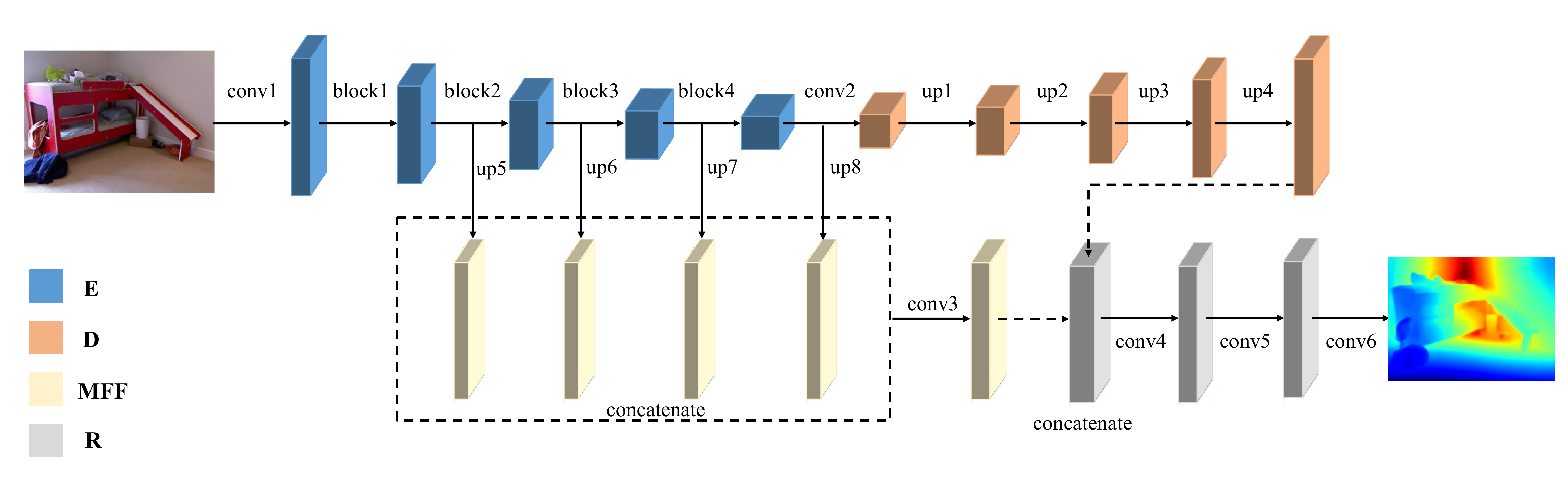}}
\caption{A diagram of the proposed network architecture. Given an input image, the encoder (E) extracts multi-scale features (1/4, 1/8, 1/16, and 1/32). The decoder (D) converts the last 1/32 scale feature to get a 1/2 scale feature. Each of the multi-scale features is up-scaled to 1/2 scale, and fused by the multi-scale feature fusion module (MFF). The outputs of D and MFF and are refined by the refinement module (R) to obtain the final depth map. Each box named ``block$n$'' denotes a block of multiple convolutional layers, such as residual block of ResNet;
each box named ``up$n$'' denotes a up-projection layer introduced in \cite{laina2016deeper}. Batch normalization and ReLU nonlinearity are applied to the output of each convolutional layer except conv6.} 
\label{fig_arch}
\end{figure*}

To summarize, we propose an improved architectural design that consists of four modules, namely, i) an encoder (E) used for multi-scale feature extraction, ii) a decoder (D) used for feature decoding, iii) a multi-scale feature fusion (MFF) module to fuse features of different resolutions extracted by E, and iv) a refinement module (R) to refine and integrate features obtained from D and MFF for final prediction. The proposed framework is shown in Fig.~\ref{fig_arch}.
Note that the employment of symmetric skip connections was widely used in previous studies on many tasks, it needs pay attention to fix the channels of the same scale features between the encoder and decoder, while the proposed feature fusion module is more flexible to use and it can be used with different backbone networks employed for the decoder-encoder part.
In our experiments, we show results obtained with several backbone networks, such as ResNet~\cite{he2016deep}, DenseNet \cite{huang2016densely}, and SENet~\cite{hu2018senet}. 
Our method is trained in an end to end fashion without employing any post-processing refinement. 

We also propose an improved loss function for training the CNNs.
It is motivated by the previous studies on statistical properties of range images of {\em natural scenes} \cite{huang2000statistics,LeeIJCV03,kalkan}. In \cite{huang2000statistics}, the authors analyzed distribution of three-dimensional points using co-occurrence statistics, and two- and three-dimensional joint distributions of Haar filter reactions. The results indicate that the range images have much simpler structures than optical images, which is also known as the ``random collage model'', that is, the world can be broken down into piecewise smooth regions that depend little on each other and sharp discontinuities in between them.
The sharp discontinuities typically emerge at the occluding boundaries of objects in scenes, which form step edges in depth maps. This structure is a key property of depth maps (of natural scenes).
As mentioned above, however, previous methods often fail to recover such edges correctly.

We argue that distorted and blurry edges emerged in the depth maps estimated by previous methods are mostly attributable to the loss function employed by them. Many previous studies use the sum of differences in depth between an estimated depth map and
its ground truth for their loss function, with different norms, i.e., $\ell_2$ loss
\cite{Eigen2014depth,Li2015DepthAS,liu2016learning}, $\ell_1$ loss \cite{ma2017sparse,Park2017JointEO}, and a
robust $berhu$ loss \cite{laina2016deeper}).
We point out that this type of losses is insensitive to errors emerging at step edges, such as shift in their positions, and difference between sharp and blurry edges. 
We then propose to use two additional loss
functions, difference in gradients 
($l_{\rm grad}$)
and difference in normals to scene surfaces
($l_{\rm normal}$)
between an estimated map and its ground truth. 
We describe that the three loss functions are complementary with each other, and show that their combination contributes to improve
accuracy particularly around object edges. We show experimental results that confirm effectiveness of our approach.

\section{Related Work}

\paragraph{Skip connections} 
Skip connections have been widely used for many purposes. In particular, they are employed in encoder-decoder networks to recover the degraded resolution due to downsampling, as in U-Net \cite{drozdzal2016importance}.
In that case, skip connections are symmetrically inserted between the same level of layers in the encoder and decoder.
This design is applied to image restoration \cite{Mao2016ImageRU} , semantic segmentation \cite{jiang2018rednet}, and others. 
In this paper, we use skip connections in a different way. Instead of symmetric application between the encoder and decoder, we directly merge the up-scaled feature maps from the different layers of the encoder to the final decoder output. This is done by the proposed multi-scale feature fusion module.

\paragraph{Gradient-based loss} Several existing studies also employ functions of depth gradients for losses \cite{Eigen2015PredictingDS,ummenhofer2017demon}.
Most of previous methods are built upon the framework of multi-task learning, such as \cite{Eigen2015PredictingDS,wang2016surge} where semantic segmentation, surface normals etc. are simultaneously learned.
On the other hand, we are interested in the following question: how far can we go in the minimal case where only ground truth depths
are available for training? This question is important, because RGBD cameras have
advanced greatly, making them easier to use, whereas segmentation labels are costly to
obtain.
Therefore, we “revisit” this minimal
formulation. We
analyze how the three different losses (i.e., $l_{depth}$, $l_{grad}$, and $l_{normal}$) can contribute to the
improvement of accuracy. 
There are other works that use gradient-based losses including DeMon \cite{ummenhofer2017demon}, which consider different problems from single image depth estimation.

\begin{figure}[ht]
\centering
\subfigure {\includegraphics[width=80mm]{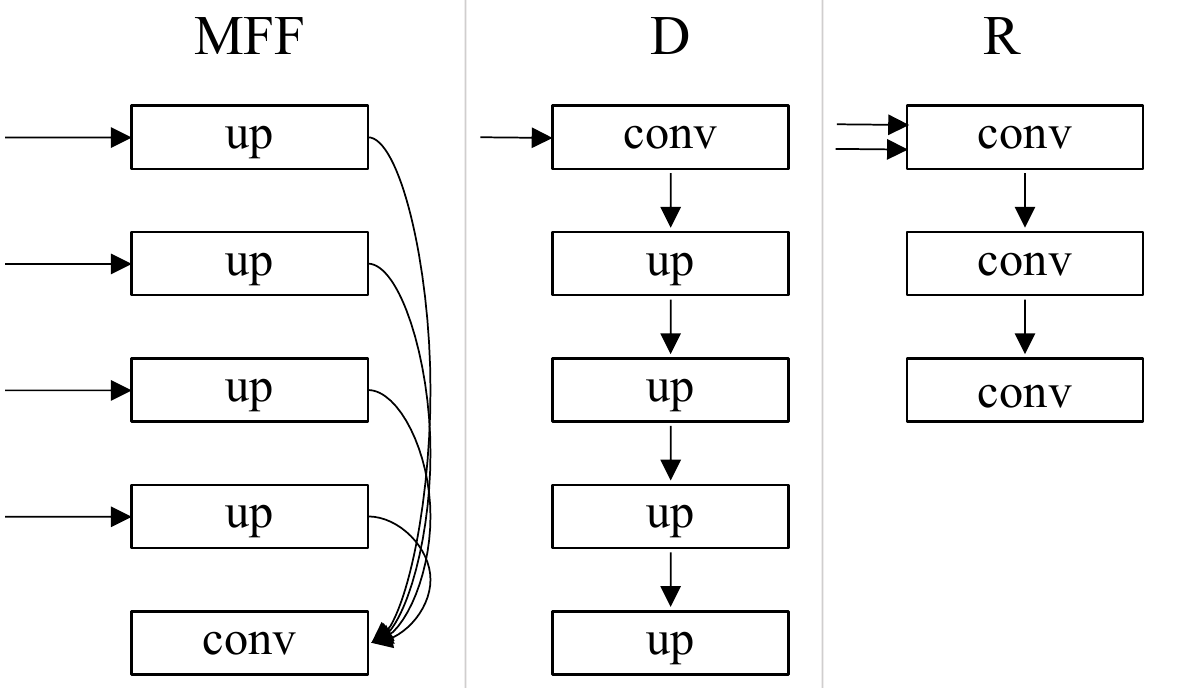}}
\caption{Diagrams of MFF, D and R. We employ the same up-sampling strategy (up-projection) used in \cite{laina2016deeper}.
} 
\label{fig_module}
\end{figure}

\begin{table}[t]
\centering  
\caption{Sizes of output features, and input/output channels of each layer when using a ResNet-50 as encoder.}
\begin{tabular}{|l|p{.1\textwidth}<{\centering}|p{.08\textwidth}<{\centering}|p{.08\textwidth}<{\centering}|} 
\hline 
Layer  &Output Size   &Input/C &Output/C \\ 
\hline\hline
conv1  &114$\times$152 &3 &64 \\
block1   &57$\times$76 &64 &256 \\
block2  &29$\times$38  &256 &512 \\  
block3  &15$\times$19  &512 &1020 \\  
block4  &8$\times$10  &1020 &2040 \\  \hline
conv2   &8$\times$10  &2040 &1020 \\  
up1   &15$\times$19 &1020 &512 \\ 
up2  &29$\times$38 &512 &256 \\ 
up3   &57$\times$76 &256 &128 \\ 
up4  &114$\times$152 &128 &64 \\  \hline
up5  &114$\times$152 &256 &16 \\ 
up6  &114$\times$152 &512 &16 \\ 
up7  &114$\times$152 &1020 &16 \\ 
up8  &114$\times$152 &2040 &16 \\ 
conv3 &114$\times$152  &64 &64 \\ \hline
conv4 &114$\times$152  &128 &128 \\
conv5 &114$\times$152  &128 &128\\
conv6 &114$\times$152  &128 &1 \\ \hline
\end{tabular}
\label{table_arch}
\end{table}

\section{Proposed Method}
\subsection{Improved Network Design}

The proposed network architecture consists of four modules: an encoder (E), a decoder (D), a multi-scale feature fusion module (MFF), and a refinement module (R), as shown in Fig.~\ref{fig_arch}. The encoder extracts features at multiple scales: 1/4, 1/8, 1/16, and 1/32. 
The decoder employs four up-projection modules \cite{laina2016deeper} to gradually up-scale the final feature from the encoder while decreasing the number of channels. Similar encoder-decoder networks were employed in \cite{laina2016deeper,ma2017sparse}, but
their outputs tend to lose spatial resolution and not to preserve object shapes. 

\begin{figure}[!t]
\centering
\subfigure[] {\includegraphics[angle=0, height=1.15in]{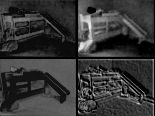}}
~~\subfigure []{\includegraphics[angle=0, height=1.15in]{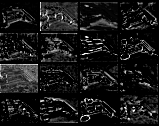}}
\caption{Visualization of outputs of different layers of the encoder network for the input image shown in Fig.~\ref{fig_arch}. Selected channels of (a)  block1, and (b) block2.}
\label{fig_feature}
\end{figure}

The MFF module integrates four different scale features from the encoder using up-projection and channel-wise concatenation. 
To be specific, the outputs of the four encoder blocks (each having 16 channels) are upsampled by $\times 2$, $4$, $8$, and $16$, respectively, so as to have the same size as the final output. This upsampling is done in a channel-wise manner. Then they are concatenated and further transformed by a convolutional layer to obtain the output of the MFF module, which has 64 channels. The main purpose of the MFF module is to merge different information at multiple scales into one. 
It is seen in Fig.~\ref{fig_feature} that the lower layer outputs of the encoder retain information with finer spatial resolution, which should be utilized to restore the fine details lost due to the multiple applications of downsampling. 

The feature from the decoder and fused multi-scale features from MFF are then concatenated in their channels and fed to the refinement module having three convolutional layers to give the final prediction. MFF and the refinement module only add a small number of parameters as compared with the encoder-decoder part. For instance, if we use the ResNet-50 as the encoder, then the encoder-decoder part has 63.6M parameters, while the feature fusion and refinement modules have only 4M parameters. 

\subsection{Loss Functions}
\label{sec:losses}

Most of the previous studies employ 
the sum of the difference between the depth estimate $d_i$ and its ground truth $g_i$ for a loss: 
\begin{equation}
    l_{\rm 1}=\frac{1}{n}\sum_{i=1}^n e_i,
\end{equation}
where $e_i = \|d_i - g_i\|_1$.
Some use its $\ell_2$ norm or robustified version. 
There are two issues with this type of losses. One is that a unit depth difference (e.g., 1cm) has an equal contribution to the loss between distant and nearby points in a scene. It should be more on nearby points and less on distant points. This is pointed out in \cite{lee2018single}, where a depth-balanced Euclidean loss is employed.
We propose a simpler remedy here, which is to use 
the logarithm of depth errors as 
\begin{equation}
    l_{\rm depth}=\frac{1}{n}\sum_{i=1}^n F(e_i),
\end{equation}
where
\begin{equation}
    F(x) = \ln(x+\alpha),
\end{equation}
where $\alpha( >0)$ is a parameter we set. 

\begin{figure}[t]
\centering
\includegraphics[width=80mm]{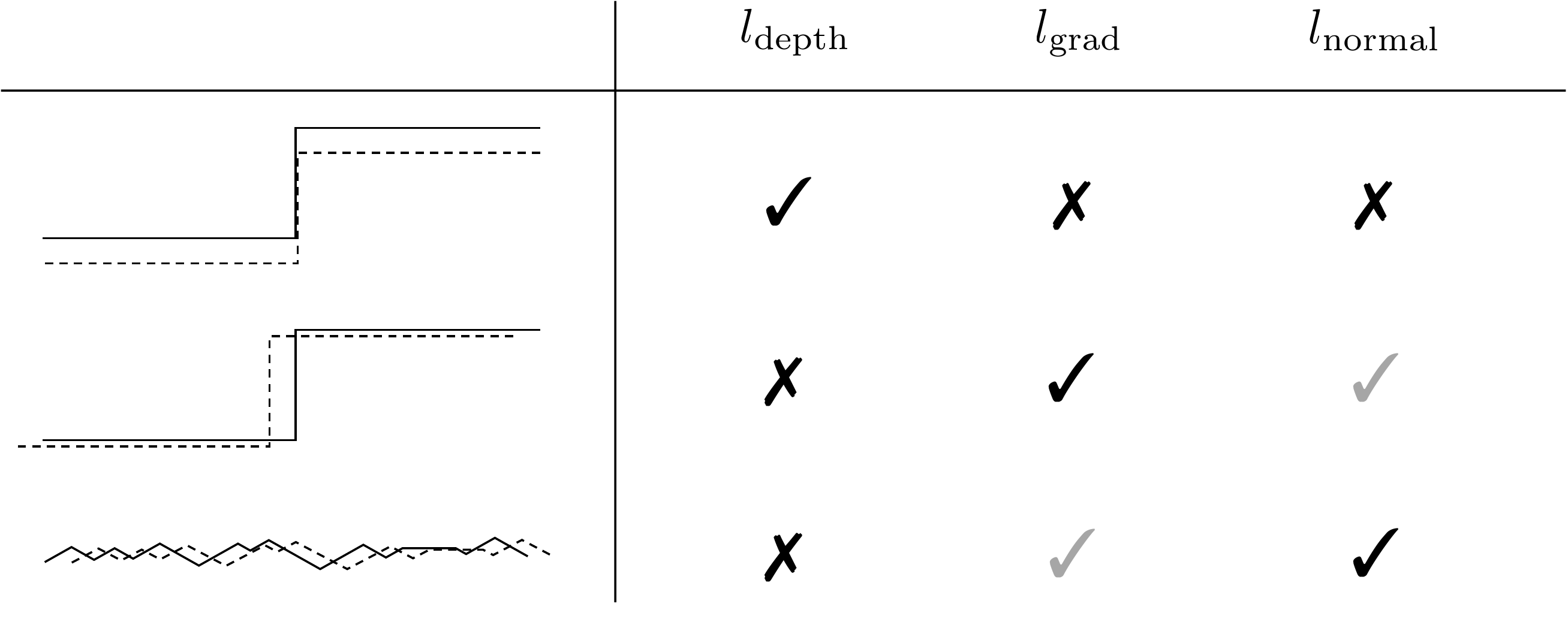}
\caption{The three loss functions have orthogonal sensitivities to different types of errors of estimated depth maps. The solid and dotted lines depicted in the first column indicate two depth maps under comparison, where they are represented by one-dimensional depth images for the sake of explanation, and the vertical axis is depth and the horizontal axis is, say, the $x$ axis of the images.}
\label{fig:losses}
\end{figure}

The other issue, which is the main concern here, is that for a step edge structure of depth, while the above conventional loss is sensitive to shifts in depth direction, it is comparatively insensitive to shifts in $x$ and $y$ directions, as illustrated in the top and second rows of Fig.~\ref{fig:losses}. 
It is similarly insensitive also to distortion and blur of edges. 

The statistics of natural range images indicate that natural scenes consist of a lot of such step edge structures \cite{huang2000statistics}, which can easily be confirmed from examples of ground truth depth maps in various datasets. 
We think that this insensitivity to small errors around edges must be a major reason for the phenomenon that the edges in depth maps estimated by CNNs trained using this loss tend to be distorted or blurry.

Thus, it is necessary to penalize such errors around edges more. Thus, we consider the following loss function of the gradients of depth;
\begin{equation}
    l_{\rm grad}=\frac{1}{n}\sum_{i=1}^n (F(\nabla_{x}(e_i))+F(\nabla_{y}(e_i))),
\end{equation}
where $\nabla_{x}(e_{i})$ is the spatial derivative of $e_{i}$ computed at the $i^{th}$ pixel with respect to $x$, and so on. This loss is sensitive to the shift of edges in $x$ and $y$ directions, as shown in Fig.~\ref{fig:losses}. Note that the proposed two loss functions $l_{\rm depth}$ and $l_{\rm grad}$ work in a complementary manner for different types of errors. Thus, we use the (weighted) sum of $l_{\rm depth}$ and $l_{\rm grad}$ to train our networks. 

Depth maps of natural scenes can roughly be modeled by a limited number of smooth surfaces and step edges in between them, according to the statistics of natural range images \cite{huang2000statistics}. For instance, depth will often be discontinuous at the boundary of an object. Errors around such strong edges are well penalized by $l_{\rm grad}$. However, since depth differences at such occluding boundaries of objects can sometimes be very large, we must choose a modest (i.e., not very large) weight $\lambda >0$ on $l_{\rm grad}$ (Note that $l_{\rm grad}$ is not upper bounded.). Then, the term $\lambda l_{\rm grad}$ cannot penalize small structural errors such as those of high-frequency undulation of a surface, as shown in the bottom row of Fig.~\ref{fig:losses}.

To deal with such small depth structures and further improve fine details of depth maps, we consider yet another loss for training, which measures accuracy of the normal to the surface of an estimated depth map with respect to its ground truth. Denoting the surface normal of an estimated depth map and its ground truth by $n^d_i\equiv [-\nabla_{x}(d_{i}), -\nabla_{y}(d_{i}),1]^\top$ and ${n^g_i\equiv [-\nabla_{x}(g_{i}), -\nabla_{y}(g_{i}),1]^\top}$
respectively, we define the following loss measuring the difference between the two normals by
\begin{equation}
    l_{\rm normal} = \frac{1}{n}\sum_{i=1}^n
    \left(
        1-\frac{\langle n^d_i,n^g_i \rangle}{
            \sqrt{\langle n^d_i,n^d_i \rangle}
            \sqrt{\langle n^g_i,n^g_i \rangle}
        }
    \right),
\end{equation}
where $\langle\cdot,\cdot\rangle$ denotes the inner product of vectors. 
Although this loss is also  computed from depth gradients, it measures the angle between two surface normals, and thus is sensitive to small depth structures illustrated in the bottom row of Fig.~\ref{fig:losses}. 
Thus, we can say that $l_{\rm normal}$ is also complementary with the other two losses. Finally, we define the loss by
\begin{equation}
    L = l_{\rm depth} +  \lambda  l_{\rm grad} + \mu l_{\rm normal},
\end{equation}
where $\lambda, \mu \in \mathbb{R}$ are weighting coefficients. 

\subsection{Accuracy Measures for Depth Estimation}
Denoting the total number of (valid) pixels used in all evaluated images by $T$, we use the following accuracy measures that are commonly employed in the previous studies:
\begin{itemize} 
 \item Root mean squared error (RMS):~ $\sqrt{\frac{1}{T}\sum \limits_{i=1}^{T} (d_{i} - g_{i})^{2}}$.
 \item Mean relative error (REL):~ $\frac{1}{T}\sum \limits_{i=1}^{T} \frac{\|d_{i} - g_{i}\|_1}{g_{i}}$.
 \item Mean $\log10$ error ($\log10$):~ $\frac{1}{T}\sum \limits_{i=1}^{T} {\|\log_{10} d_{i} - \log_{10} g_{i}\|_1}$.
 \item Thresholded accuracy:~ Percentage of $d_{i}$, such that ${\max\left(\frac{d_{i}}{g_{i}},\frac{g_{i}}{d_{i}}\right) = \delta < \mbox{threshold}}$.
\end{itemize}

These popular measures enable us to compare methods from multiple aspects to evaluate depth accuracy. However, a combination of these measures still has limitation. We argue that these measures are not good at detecting spatial distortion of object edges, because of the same reason as we discussed before. For instance, the method of Ma and Karaman \cite{ma2017sparse}, which leverages known depths at a few scene points to improve depth estimation, does outperform other methods that do not use such additional information, if we use the above measures.
However, the outputs of their method tend to have spatially distorted or blurry object edges; local structures are often missing. The same tendency can be observed for others, particularly the recent ones; estimated depth maps showing small RMS errors tend to have apparent errors of this type.

\paragraph{Edge accuracy:} In order to more properly evaluate accuracy of estimated depth maps, we propose an additional measure that is sensitive to positional errors of edges which will be overlooked by the above measures. 
For this purpose, we apply the Sobel operator \cite{sobel19683x3} to both of the estimated and the true depth maps, and then apply a threshold to them to identify pixels which satisfy
$\sqrt{f_{x}(i)^{2} + f_{y}(i)^{2}} > \mbox{(threshold})$ by `pixels on edges', 
where $f_{x}$ and $f_{y}$ are $3\times3$ horizontal and vertical Sobel operators, respectively. We used three different thresholds: 0.25, 0.5, and 1. Assuming those of the true depth map to be true, we measure precision (P), recall (R) and F1 score for those of the estimated map. 

\setlength{\tabcolsep}{3.2pt}
\begin{table*}[!t]
\begin{center}
\caption{Comparisons of different methods on the NYU-Depth V2 dataset. The methods marked by $^*$ use partially known depths, and those with $^{**}$ employ joint task learning.}
\label{nyu_depth}
\begin{tabular}{|l|c|c|c|c|c|c|}
\hline
Method & RMS & REL &$\log10$& $\delta<1.25$ & $\delta<1.25^{2}$ & $\delta<1.25^{3}$\\
\hline\hline
Eigen et al. \protect\cite{Eigen2014depth} &0.907  &0.215 &- &0.611 &0.887 &0.971\\   
Liu et al. \protect\cite{liu2015deep} &0.824  &0.230 &0.095 &0.614 &0.883 &0.971   \\ 
Chakrabarti et al. \protect\cite{chakrabarti2016depth} &0.620  &0.149 &- &0.806 &0.958 &0.987\\ 
Cao et al. \protect\cite{cao2017estimating} &0.819  &0.232 &0.091 &0.646 &0.892 &0.968\\
Li et al. \protect\cite{li2017two}  &0.635 &0.143 &0.063 &0.788 &0.958 &0.991 \\
Ma and Karaman  \protect\cite{ma2017sparse} 	&(-)  &0.143 &- &0.810 &0.959 &0.989	 \\    
Laina et al. \protect\cite{laina2016deeper} &0.573 &0.127 &0.055 &0.811 &0.953 &0.988 \\ 
Xu et al. \protect\cite{Xu2017MultiscaleCC} &0.586  &0.121 &0.052 &0.811 &0.954 &0.987	\\  
Lee et al. \protect\cite{lee2018single} &0.572  &0.139 &- &0.815 &0.963 &0.991	 \\    
Fu et al. \protect\cite{fu2018deep} 	&\textbf{0.509}  &\textbf{0.115} &0.051 &0.828 &0.965 &0.992	 \\    
Qi et al. \protect\cite{qi2018geonet} &0.569  &0.128 &0.057 &0.834 &0.960 &0.990	 \\  
\hline
Ours (ResNet-50) &0.555 &0.126&0.054 &0.843 &0.968 &0.991\\ 
Ours (DenseNet-161) &0.544 &0.123 &0.053 &0.855 &0.972 &0.993\\ 
Ours (SENet-154) &0.530 &\textbf{0.115}  &\textbf{0.050} &\textbf{0.866} &\textbf{0.975} &\textbf{0.993}\\\hline
Ma and Karaman \protect\cite{ma2017sparse}$^{*}$ 	&(-)   &0.044&- &0.971 &0.994 &0.998	 \\  
Li et al. \protect\cite{Li2015DepthAS} &0.821  &0.232 &0.094 &0.621 &0.886 &0.968  \\ 
Eigen and Fergus \protect\cite{Eigen2015PredictingDS}$^{**}$ & 0.641  &0.158&- & 0.769 & 0.950 & 0.988	\\ 
Dharmasiri et al. \protect\cite{Dharmasiri2017JointPO}$^{**}$	&0.624  &0.156&- &0.776 &0.953 &0.989	\\  
Xu et al. \protect\cite{xu2018pad}$^{**}$	&0.582  &0.120&0.055 &0.817 &0.954 &0.987	\\  
\hline
\end{tabular}
\end{center}
\end{table*}
\setlength{\tabcolsep}{1.4pt}

\setlength{\tabcolsep}{3.4pt}
\begin{table}[!t]
\begin{center}
\caption{Edge accuracy for the NYU-Depth V2. See text for details. The method marked by $^{**}$ employs joint task learning.
}\label{nyu_shape}
\begin{tabular}{|l|l|c|c|c|}
\hline
Threshold &Method&P & R &F1 \\
\hline\hline
$>$0.25 
&Eigen and Fergus \protect\cite{Eigen2015PredictingDS}$^{**}$ &0.544&0.481&0.500\\
&Laina et al \protect\cite{laina2016deeper}  &0.489 &0.435 &0.454 \\
&Xu et al. \cite{Xu2017MultiscaleCC} &0.516 &0.400 &0.436 \\
&Fu et al. \cite{fu2018deep} &0.320 &0.583 &0.402 \\ 
&Ours (SENet-154)   &0.644 &0.508 &0.562 \\
\hline
$>$0.5
&Eigen and Fergus \protect\cite{Eigen2015PredictingDS}$^{**}$ &0.587&0.456&0.501\\
&Laina et al \protect\cite{laina2016deeper}&0.536&0.422&0.463\\
&Xu et al. \cite{Xu2017MultiscaleCC} &0.600 &0.366 &0.439 \\
&Fu et al. \cite{fu2018deep} &0.316 &0.473 &0.412 \\
&Ours (SENet-154)       &0.668 &0.505 &0.568 \\
\hline
$>$1
&Eigen and Fergus \protect\cite{Eigen2015PredictingDS}$^{**}$ &0.733&0.488&0.574\\
&Laina et al \protect\cite{laina2016deeper}&0.670&0.479&0.548\\
&Xu et al. \cite{Xu2017MultiscaleCC} &0.794 &0.407 &0.525 \\
&Fu et al. \cite{fu2018deep} &0.483 &0.512 &0.485 \\
&Ours (SENet-154)        &0.759 &0.540 &0.623 \\
\hline
\end{tabular}
\end{center}
\end{table}
\setlength{\tabcolsep}{1.4pt}

\section{Experiments}

\begin{figure*}[!t]
\centering  
\begin{tabular}{ccccccc}
\IncG[ width=0.9in]{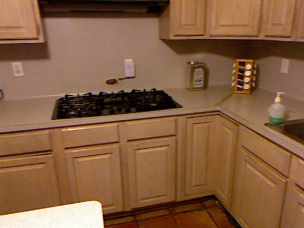}
&\IncG[ width=0.9in]{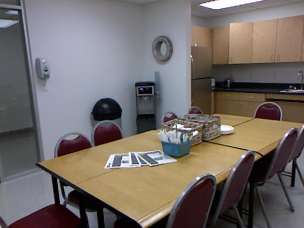}
&\IncG[ width=0.9in]{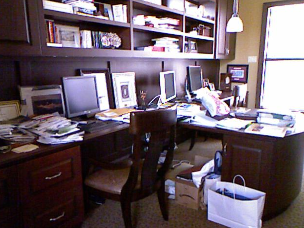}
&\IncG[ width=0.9in]{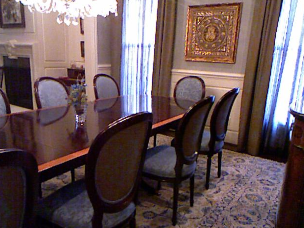}
&\IncG[ width=0.9in]{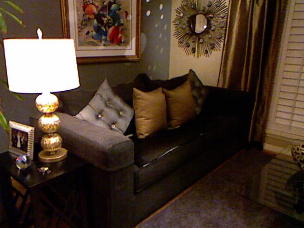}
&\IncG[ width=0.9in]{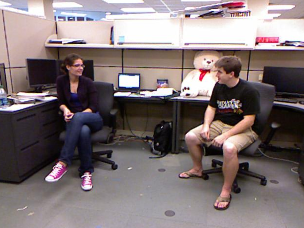}
&RGB\\
\IncG[ width=0.9in]{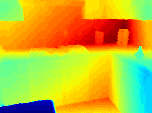}
&\IncG[ width=0.9in]{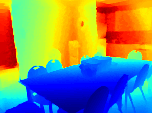}
&\IncG[ width=0.9in]{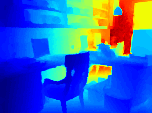}
&\IncG[ width=0.9in]{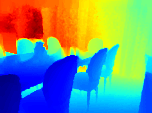}
&\IncG[ width=0.9in]{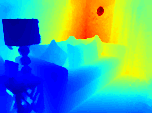}
&\IncG[ width=0.9in]{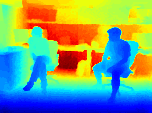}
&\begin{minipage}{.12\textwidth}\centering
\setlength{\baselineskip}{1.0em}Ground truth\end{minipage}\\
\IncG[ width=0.9in]{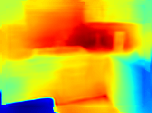}
&\IncG[ width=0.9in]{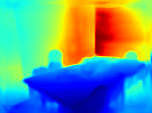}
&\IncG[ width=0.9in]{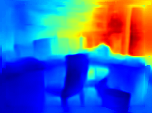}
&\IncG[ width=0.9in]{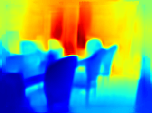}
&\IncG[ width=0.9in]{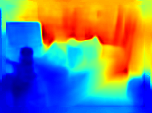}
&\IncG[ width=0.9in]{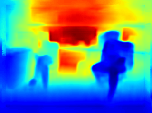}
&\begin{minipage}{.12\textwidth}\centering
\setlength{\baselineskip}{1.0em}
Eigen and Fergus \protect\cite{Eigen2015PredictingDS}$^{**}$\end{minipage} \\
\IncG[ width=0.9in]{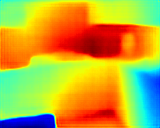}
&\IncG[ width=0.9in]{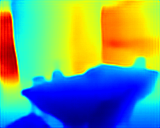}
&\IncG[ width=0.9in]{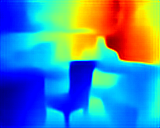}
&\IncG[ width=0.9in]{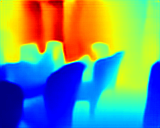}
&\IncG[ width=0.9in]{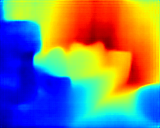}
&\IncG[ width=0.9in]{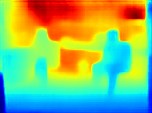}
&\begin{minipage}{.12\textwidth}\centering
\setlength{\baselineskip}{1.0em}
Laina et al. \protect\cite{laina2016deeper}\end{minipage}\\
\IncG[ width=0.9in]{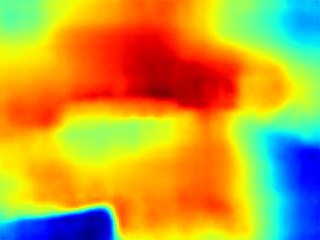}
&\IncG[ width=0.9in]{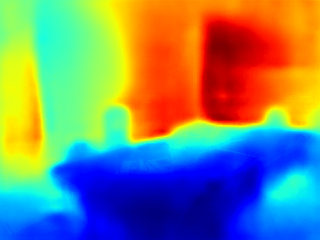}
&\IncG[ width=0.9in]{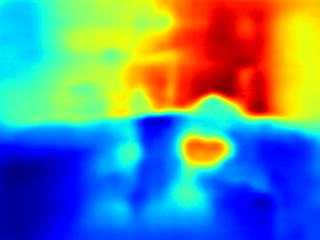}
&\IncG[ width=0.9in]{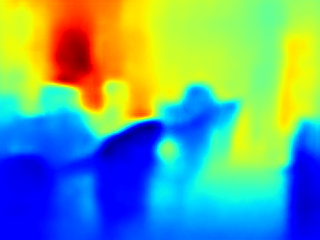}
&\IncG[ width=0.9in]{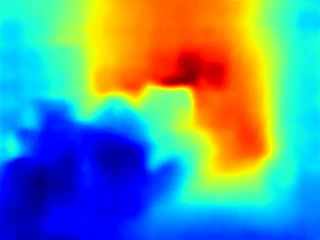}
&\IncG[ width=0.9in]{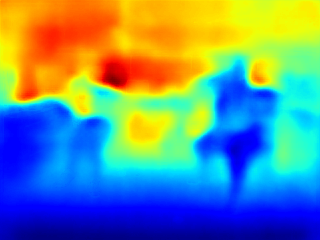}
&\begin{minipage}{.12\textwidth}\centering
\setlength{\baselineskip}{1.0em}
Xu et al. \protect\cite{Xu2017MultiscaleCC}\end{minipage}\\
\IncG[ width=0.9in]{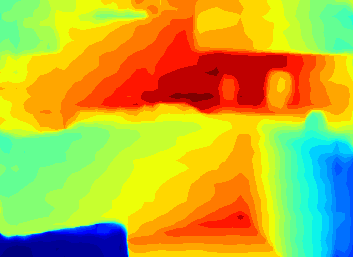}
&\IncG[ width=0.9in]{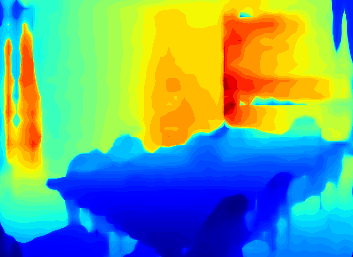}
&\IncG[ width=0.9in]{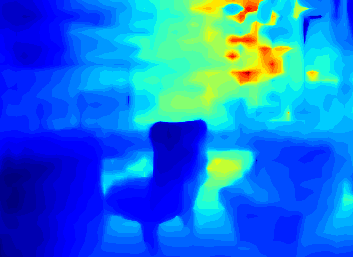}
&\IncG[ width=0.9in]{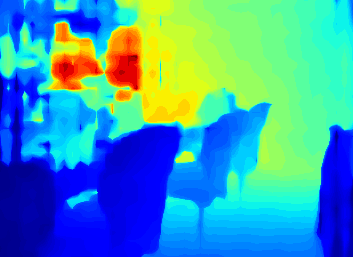}
&\IncG[ width=0.9in]{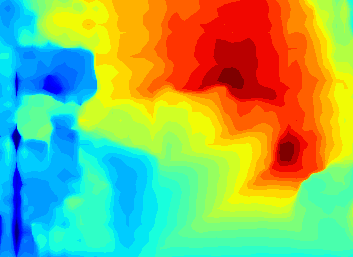}
&\IncG[ width=0.9in]{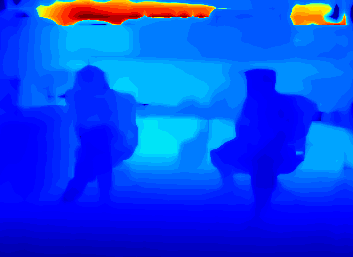}
&\begin{minipage}{.12\textwidth}\centering
\setlength{\baselineskip}{1.0em}
Fu et al. \protect\cite{fu2018deep}\end{minipage} \\
\IncG[ width=0.9in]{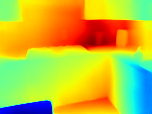}
&\IncG[ width=0.9in]{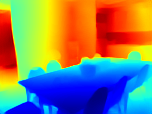}
&\IncG[ width=0.9in]{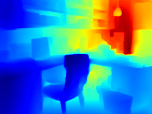}
&\IncG[ width=0.9in]{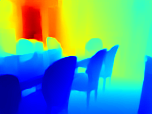}
&\IncG[ width=0.9in]{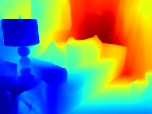}
&\IncG[ width=0.9in]{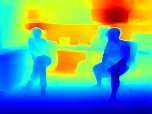}
&\begin{minipage}{.12\textwidth}\centering
\setlength{\baselineskip}{1.0em} 
Ours \\
 \end{minipage}
\end{tabular}
\caption{Results of different methods for six images. From the first to the last row; input RGB images, ground truth depth map, a multi-task learning method
\cite{Eigen2015PredictingDS}, encoder-decoder network \cite{laina2016deeper}, CRF-based method
\cite{Xu2017MultiscaleCC}, 
dilated ordinary regression network
\cite{fu2018deep}, and our proposed network trained with the full loss function. We show them in the ascending order of quality in traditional measures. }
\label{fig_nyu}
\end{figure*}

\subsection{Implementation Details}
We use the NYU-Depth V2 dataset \cite{Silberman2012IndoorSA} which consists of a variety of indoor scenes, and is the most widely used for the task of single view depth prediction.
In the previous studies \cite{Eigen2014depth,laina2016deeper,ma2017sparse,Eigen2015PredictingDS,Xu2017MultiscaleCC}, it is shown that data argumentation helps to improve accuracy as well as to avoid over-fitting. We employ the following data augmentation methods, which are individually applied to each sample (an RGB image and the corresponding depth map) in an online manner:
\begin{itemize}
 \item Flip: The RGB and the depth image are both horizontally flipped with 0.5 probability.
 \item Rotation: The RGB and the depth image are both rotated by a random degree ${r\in[-5,5]}$.
 \item Color Jitter: Brightness, contrast, and saturation values of the RGB image are randomly scaled by $c\in[0.6,1.4]$.
\end{itemize}

We follow the same procedure as the one employed in the previous studies.
We use the official splits for 464 scenes, i.e., 249 scenes for training and 215 scenes for testing. 
Following previous methods, we downsample images from original size (640$\times$480) to 320$\times$240 pixels using bilinear interpolation, and then crop their central parts to obtain images with 304$\times$228 pixels.  For training, the depth maps are downsampled to 114$\times$152 to fit the size of output. 
For testing, following the previous studies, we use the same small subset of 654 samples. 

We train the proposed network for 20 epochs. The encoder module in the network is initialized by a model pretained with the ImageNet dataset \cite{deng2009imagenet}. The other layers in the network are randomly initialized. 
We use Adam optimizer with an initial learning rate of 0.0001, and reduce it to 10$\%$ for every 5 epochs. We set $\beta_{1}=0.9$, $\beta_{2}=0.999$, and use weight decay of 0.0001.
The weights $\lambda$ of $l_{\rm grad}$ and $\mu$ of $l_{\rm normal}$ are set as $\lambda=1$ and $\mu=1$, and $\alpha$ in the mapping function is set to 0.5 in all the experiments. 
We conducted all the experiments using PyTorch \cite{paszke2017automatic} with batch size of 8.

\subsection{Performance Comparison}
Table~\ref{nyu_depth}\footnote{RMS values given in \cite{ma2017sparse} are omitted here (displayed as (-)), because we found an error in the authors' code, and believe that the numbers reported in their paper are miscalculated.} shows the results of our method together with those of existing methods on the three basic measures. For the sake of references, the table includes the methods  that use additional information other than the input RGB images; those with $^*$ use relative depth between pairs or partially known depths \cite{chen2016single,ma2017sparse},
and those with  $^{**}$ employ joint task learning \cite{ladicky2014pulling,wang2015towards,Li2015DepthAS,Eigen2015PredictingDS,Dharmasiri2017JointPO,xu2018pad}. For the method of Ma and Karaman \cite{ma2017sparse}, we show two results that are obtained from single RGB images alone and with partially known depths (200 pixels). The methods denoted without a superscript \cite{Eigen2014depth,laina2016deeper,Xu2017MultiscaleCC,liu2015deep,cao2017estimating,chakrabarti2016depth,li2017two,lee2018single,qi2018geonet,fu2018deep} and ours should be able to be compared in an equal condition. 

 Our method achieved the best performance on REL, $\log10$, and outperformed previous methods for
$\delta<1.25$, $\delta<1.25^{2}$,  $\delta<1.25^{3}$ with a large margin.
It is observed that our method provided the second best performance for RMS, while the method of \cite{fu2018deep} performed a little better on this measure. However, they used 120K training samples, and we used less number of training samples (50K samples).
In addition, our method outperformed the methods proposed in \cite{laina2016deeper,ma2017sparse}, which employed encoder-decoder networks with ResNet-50. 

Figure \ref{fig_nyu} shows the depth maps estimated by different methods\footnote{For \cite{Eigen2015PredictingDS}, we show the results that are made publicly available by the authors. For \cite{laina2016deeper,Xu2017MultiscaleCC,fu2018deep}, we use the authors' code to obtain the results and show them here. }, including joint multi-task learning method \cite{Eigen2015PredictingDS}, encoder-decoder network \cite{laina2016deeper}, CRF-based refinement method \cite{Xu2017MultiscaleCC}, dilated ordinary regression network \cite{fu2018deep} and our method.
We show them in the ascending order of quality in traditional measures.
It is seen that the methods of \cite{laina2016deeper,Xu2017MultiscaleCC} suffer from heavy distortion of shapes, and the method of \cite{fu2018deep} produces sharp discontinuities of shapes of objects, and mosaic effect is even observed.
Although the method of \cite{Eigen2015PredictingDS} shows relatively clear image boundaries, they produced many inaccurate weak edges which can be observed in the far side of each room. 
Our method shows significant performance by correctly recovering edges of objects  and small structures, such as bottles in a kitchen and lamp shades on a desk.

We also observe a relationship between our proposed measure of edge accuracy   and visual quality of estimated depth maps. For instance, our proposed method outperforms all the other methods given in Table~\ref{nyu_shape}, and provides depth maps with the finest details in Fig.\ref{fig_nyu}. A similar relationship is also observed for the method of \cite{Eigen2015PredictingDS} which shows the best scores among the previous works in Table~\ref{nyu_shape}, and provides better visual results compared to these related work in Fig.\ref{fig_nyu}.   

\setlength{\tabcolsep}{4pt}
\begin{table}[!t]
\begin{center}
\caption{Results of our method that is built on ResNet-50 trained with different loss functions on the NYU-Depth V2 dataset. For edge accuracy, we report the results for $>$0.5.}
\label{results_loss}
\begin{tabular}{|l|c|c|c|c|}
\hline
 & RMS & REL & $\delta<1.25$ &F1 \\
\hline\hline
w/~ $l_{\rm depth}$ &0.580  &0.133 &0.830 &0.525\\ 
w/~ $l_{\rm depth}+\lambda l_{\rm grad}$  &0.563  &0.128 &0.841 &0.543\\ 
w/~ full loss &\textbf{0.555}  &\textbf{0.126} &\textbf{0.843} &\textbf{0.548}\\
\hline
\end{tabular}
\end{center}
\end{table}
\setlength{\tabcolsep}{1.4pt}

\begin{figure}[!t]
\centering  
\begin{tabular}{ccc}
\IncG[height=28mm]{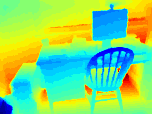}
&~\IncG[height=28mm]{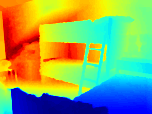}
\\
\IncG[height=28mm]{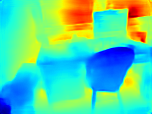}
&~\IncG[height=28mm]{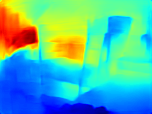}
\\
\IncG[height=28mm]{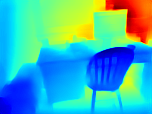}
&~\IncG[height=28mm]{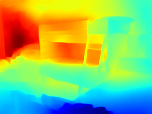}
\end{tabular}
\caption{Visual comparison of our method trained using ResNet-50  with different loss functions on the NYU-Depth V2 dataset. From top to bottom: Ground Truth, trained with $l_{depth}$ and the full loss, respectively.}
\label{fig_nyu_loss}
\end{figure}

\subsection{Ablation Study}

In order to compare and analyze the performance of the proposed loss functions, we train our model with different losses while using ResNet-50 as an encoder. 
In visual comparison of their estimated depth maps side by side, as  shown in Figure~\ref{fig_nyu_loss} with several samples, it can be observed that the results trained with $l_{\rm depth}$ are more distorted and blurry.
Moreover, the proposed loss function provides significant improvement especially for estimation of fine details and clear boundaries of objects in scenes. The numerical results given in Table~\ref{results_loss}, show that finer details of objects in the scenes are gradually recovered using a model with $l_{\rm depth}$, $l_{\rm depth} + \lambda l_{\rm grad}$ and the full loss $l_{\rm depth} + \lambda l_{\rm grad} + \mu l_{\rm normal}$, respectively.

\section{Conclusion}

 

We have presented two improvements to existing methods for single image depth estimation. One is the improved design of network architecture. It consists of four modules: an encoder, a decoder, a multi-scale feature fusion module and a refinement module. Any base network can be used for the encoder,  
such as ResNet, DenseNet, and SENet. The overall network is trained in an end to end
fashion without any post-processing refinement. 

The previous methods fail to correctly estimate boundaries of objects
in scenes. We explain that this failure may be attributable to the loss functions employed by the previous
methods. The analyses of natural range image statistics indicate that the real world scene can be decomposed into smooth surfaces and sharp discontinuities in between them; the latter
corresponds to object boundaries. Then, this makes it important to be able to accurately reconstruct those discontinuities, which appear as step edges in depth maps, and to deal
with them appropriately during training of CNNs. We have
made a simple analysis of how different loss functions affect
measurement of estimation errors around step edges. Based
on it, we argue that the loss of difference in depth is insensitive to positional shift and blurring of the edges, whereas the
loss of difference in gradients tends to be sensitive to them.
We further employ an additional loss of difference in surface
normals, which is expected to be sensitive to small structures
that tend to be neglected by the above two losses. We then
propose to use a combined loss of the three loss functions.

Finally, we presented experimental results on the NYU Depth V2 dataset. We observed that existing measures, which make use of difference
in depth, fail to correctly measure reconstruction
error of step edges for evaluation of estimation accuracy. For more proper evaluation, we presented a simple measure of reconstruction accuracy of step
edges. Our method outperforms previous methods using both
traditional measures and the proposed measure, especially providing
remarkable improvement on reconstruction of small objects and estimation of object boundaries. This agrees well with visual comparisons between the results of the proposed method and those of the previous ones.

{\small
\bibliographystyle{ieee}
\bibliography{egbib}

\begin{thebibliography}{10}\itemsep=-1pt

\bibitem{LeeIJCV03}
L.~A. B, P.~K. S, and M.~David.
\newblock The nonlinear statistics of high-contrast patches in natural images.
\newblock {\em IJCV}, 54(1-3):83--103, 2003.

\bibitem{cao2017estimating}
Y.~Cao, Z.~Wu, and C.~Shen.
\newblock Estimating depth from monocular images as classification using deep
  fully convolutional residual networks.
\newblock {\em IEEE Transactions on Circuits and Systems for Video Technology},
  2017.

\bibitem{chakrabarti2016depth}
A.~Chakrabarti, J.~Shao, and G.~Shakhnarovich.
\newblock Depth from a single image by harmonizing overcomplete local network
  predictions.
\newblock In {\em NIPS}, pages 2658--2666, 2016.

\bibitem{chen2016single}
W.~Chen, Z.~Fu, D.~Yang, and J.~Deng.
\newblock Single-image depth perception in the wild.
\newblock In {\em NIPS}, pages 730--738, 2016.

\bibitem{deng2009imagenet}
J.~Deng, W.~Dong, R.~Socher, L.-J. Li, K.~Li, and L.~Fei-Fei.
\newblock Imagenet: A large-scale hierarchical image database.
\newblock In {\em CVPR}, pages 248--255, 2009.

\bibitem{Dharmasiri2017JointPO}
T.~Dharmasiri, A.~Spek, and T.~Drummond.
\newblock Joint prediction of depths, normals and surface curvature from rgb
  images using cnns.
\newblock {\em CoRR}, abs/1706.07593, 2017.

\bibitem{drozdzal2016importance}
M.~Drozdzal, E.~Vorontsov, G.~Chartrand, S.~Kadoury, and C.~Pal.
\newblock The importance of skip connections in biomedical image segmentation.
\newblock In {\em Deep Learning and Data Labeling for Medical Applications},
  pages 179--187. Springer, 2016.

\bibitem{Eigen2015PredictingDS}
D.~Eigen and R.~Fergus.
\newblock Predicting depth, surface normals and semantic labels with a common
  multi-scale convolutional architecture.
\newblock {\em ICCV}, pages 2650--2658, 2015.

\bibitem{Eigen2014depth}
D.~Eigen, C.~Puhrsch, and R.~Fergus.
\newblock Depth map prediction from a single image using a multi-scale deep
  network.
\newblock In {\em NIPS}, pages 2366--2374, 2014.

\bibitem{fu2018deep}
H.~Fu, M.~Gong, C.~Wang, K.~Batmanghelich, and D.~Tao.
\newblock Deep ordinal regression network for monocular depth estimation.
\newblock In {\em CVPR}, pages 2002--2011, 2018.

\bibitem{huang2016densely}
H.~Gao, L.~Zhuang, W.~K. Q, and van~der Maaten~Laurens.
\newblock Densely connected convolutional networks.
\newblock {\em CVPR}, 2017.

\bibitem{Gupta2013HumanAR}
R.~K. Gupta, A.~Y.~S. Chia, and D.~Rajan.
\newblock Human activities recognition using depth images.
\newblock In {\em ACM Multimedia}, 2013.

\bibitem{he2016deep}
K.~He, X.~Zhang, S.~Ren, and J.~Sun.
\newblock Deep residual learning for image recognition.
\newblock In {\em CVPR}, pages 770--778, 2016.

\bibitem{hu2018senet}
J.~Hu, L.~Shen, and G.~Sun.
\newblock Squeeze-and-excitation networks.
\newblock In {\em CVPR}, 2018.

\bibitem{huang2000statistics}
J.~Huang, A.~B. Lee, and D.~Mumford.
\newblock Statistics of range images.
\newblock In {\em CVPR}, volume~1, pages 324--331, 2000.

\bibitem{laina2016deeper}
L.~Iro, R.~Christian, B.~Vasileios, T.~Federico, and N.~Nassir.
\newblock Deeper depth prediction with fully convolutional residual networks.
\newblock In {\em 3DV}, pages 239--248, 2016.

\bibitem{sobel19683x3}
S.~Irwin and F.~Gary.
\newblock A 3x3 isotropic gradient operator for image processing.
\newblock {\em a talk at the Stanford Artificial Project in}, pages 271--272,
  1968.

\bibitem{jiang2018rednet}
J.~Jiang, L.~Zheng, F.~Luo, and Z.~Zhang.
\newblock Rednet: Residual encoder-decoder network for indoor rgb-d semantic
  segmentation.
\newblock {\em arXiv preprint arXiv:1806.01054}, 2018.

\bibitem{kalkan}
S.~Kalkan, F.~W{\"o}rg{\"o}tter, and N.~Kr{\"u}ger.
\newblock Statistical analysis of local 3d structure in 2d images.
\newblock In {\em CVPR}, volume~1, pages 1114--1121, June 2006.

\bibitem{ladicky2014pulling}
L.~Ladicky, J.~Shi, and M.~Pollefeys.
\newblock Pulling things out of perspective.
\newblock In {\em CVPR}, pages 89--96, 2014.

\bibitem{lee2018single}
J.-H. Lee, M.~Heo, K.-R. Kim, and C.-S. Kim.
\newblock Single-image depth estimation based on fourier domain analysis.
\newblock In {\em CVPR}, pages 330--339, 2018.

\bibitem{lee2011depth}
W.~Lee, N.~Park, and W.~Woo.
\newblock Depth-assisted real-time 3d object detection for augmented reality.
\newblock {\em ICAT}, 2:126--132, 2011.

\bibitem{Li2015DepthAS}
B.~Li, C.~Shen, Y.~Dai, A.~van~den Hengel, and M.~He.
\newblock Depth and surface normal estimation from monocular images using
  regression on deep features and hierarchical crfs.
\newblock {\em CVPR}, pages 1119--1127, 2015.

\bibitem{li2017two}
J.~Li, R.~Klein, and A.~Yao.
\newblock A two-streamed network for estimating fine-scaled depth maps from
  single rgb images.
\newblock In {\em CVPR}, pages 3372--3380, 2017.

\bibitem{liu2015deep}
F.~Liu, C.~Shen, and G.~Lin.
\newblock Deep convolutional neural fields for depth estimation from a single
  image.
\newblock In {\em CVPR}, pages 5162--5170, 2015.

\bibitem{liu2016learning}
F.~Liu, C.~Shen, G.~Lin, and I.~D. Reid.
\newblock Learning depth from single monocular images using deep convolutional
  neural fields.
\newblock {\em PAMI}, 38(10):2024--2039, 2016.

\bibitem{Lu2018CurveStructureSF}
Y.~Lu, J.~Zhou, J.~Wang, J.~Chen, K.~Smith, C.~Wilder, and S.~Wang.
\newblock Curve-structure segmentation from depth maps: A cnn-based approach
  and its application to exploring cultural heritage objects.
\newblock {\em AAAI}, 2018.

\bibitem{ma2017sparse}
F.~Ma and S.~Karaman.
\newblock Sparse-to-dense: Depth prediction from sparse depth samples and a
  single image.
\newblock {\em ICRA}, 2018.

\bibitem{Mao2016ImageRU}
X.-J. Mao, C.~Shen, and Y.-B. Yang.
\newblock Image restoration using very deep convolutional encoder-decoder
  networks with symmetric skip connections.
\newblock In {\em NIPS}, 2016.

\bibitem{Park2017JointEO}
H.~Park and K.~M. Lee.
\newblock Joint estimation of camera pose, depth, deblurring, and
  super-resolution from a blurred image sequence.
\newblock {\em ICCV}, pages 4623--4631, 2017.

\bibitem{paszke2017automatic}
A.~Paszke, S.~Gross, and A.~Lerer.
\newblock Automatic differentiation in pytorch.
\newblock 2017.

\bibitem{qi2018geonet}
X.~Qi, R.~Liao, Z.~Liu, R.~Urtasun, and J.~Jia.
\newblock Geonet: Geometric neural network for joint depth and surface normal
  estimation.
\newblock In {\em CVPR}, pages 283--291, 2018.

\bibitem{Ren2011DepthCB}
Z.~Ren, J.~Meng, and J.~Yuan.
\newblock Depth camera based hand gesture recognition and its applications in
  human-computer-interaction.
\newblock {\em 2011 8th International Conference on Information, Communications
  and Signal Processing}, pages 1--5, 2011.

\bibitem{Silberman2012IndoorSA}
N.~Silberman, D.~Hoiem, P.~Kohli, and R.~Fergus.
\newblock Indoor segmentation and support inference from rgbd images.
\newblock In {\em ECCV}, 2012.

\bibitem{song2017depth}
X.~Song, L.~Herranz, and S.~Jiang.
\newblock Depth cnns for rgb-d scene recognition: Learning from scratch better
  than transferring from rgb-cnns.
\newblock In {\em AAAI}, pages 4271--4277, 2017.

\bibitem{ummenhofer2017demon}
B.~Ummenhofer, H.~Zhou, J.~Uhrig, N.~Mayer, E.~Ilg, A.~Dosovitskiy, and
  T.~Brox.
\newblock Demon: Depth and motion network for learning monocular stereo.
\newblock In {\em CVPR}, volume~5, page~6, 2017.

\bibitem{wang2015towards}
P.~Wang, X.~Shen, Z.~L. Lin, S.~Cohen, B.~L. Price, and A.~L. Yuille.
\newblock Towards unified depth and semantic prediction from a single image.
\newblock In {\em CVPR}, pages 2800--2809, 2015.

\bibitem{wang2016surge}
P.~Wang, X.~Shen, B.~Russell, S.~Cohen, B.~L. Price, and A.~L. Yuille.
\newblock Surge: Surface regularized geometry estimation from a single image.
\newblock In {\em NIPS}, pages 172--180, 2016.

\bibitem{xu2018pad}
D.~Xu, W.~Ouyang, X.~Wang, and N.~Sebe.
\newblock Pad-net: Multi-tasks guided prediction-and-distillation network for
  simultaneous depth estimation and scene parsing.
\newblock {\em CVPR}, 2018.

\bibitem{Xu2017MultiscaleCC}
D.~Xu, E.~Ricci, W.~Ouyang, X.~Wang, and N.~Sebe.
\newblock Multi-scale continuous crfs as sequential deep networks for monocular
  depth estimation.
\newblock {\em CVPR}, pages 161--169, 2017.

\bibitem{zia2017rgb}
S.~Zia, B.~Y{\"u}ksel, D.~Y{\"u}ret, and Y.~Yemez.
\newblock Rgb-d object recognition using deep convolutional neural networks.
\newblock In {\em 2017 IEEE International Conference on Computer Vision
  Workshops (ICCVW)}, pages 887--894. IEEE, 2017.

\end{thebibliography}
}

\end{document}